# Human-Robot Creative Interactions (HRCI): Exploring Creativity in Artificial Agents Using a Story-Telling Game

Eduardo Benítez Sandoval [1,*], Ricardo Sosa [2,3], Massimiliano Cappuccio [4], and Tomasz Bednarz [1,5]

[1] Faculty of Arts, Design and Architecture, University of New South Wales, Sydney, New South Wales, Australia.
[2] Creative Technologies, Auckland University of Technology, Auckland, New Zealand.
[3] Monash Art, Design and Architecture, Monash University, Victoria, Australia.
[4] UNSW Canberra, ACT, Australia.
[5] UNSW's EPICentre and CSIRO's Data61, Sydney, Australia.

Correspondence*:
Eduardo B. Sandoval
e.sandoval@unsw.edu.au

## ABSTRACT

Creativity in social robots requires further attention in the interdisciplinary field of Human-Robot Interaction (HRI). This paper investigates the hypothesised connection between the perceived creative agency and the animacy of social robots. The goal of this work is to assess the relevance of robot movements in the attribution of creativity to robots. The results of this work inform the design of future Human-Robot Creative Interactions (HRCI). The study uses a storytelling game based on visual imagery inspired by the game 'Story Cubes' to explore the perceived creative agency of social robots. This game is used to tell a classic story for children with an alternative ending. A 2x2 experiment was designed to compare two conditions: the robot telling the original version of the story and the robot plot-twisting the end of the story. A Robotis Mini humanoid robot was used for the experiment. As a novel contribution, we propose an adaptation of the Short Scale Creative Self scale (SSCS) to measure perceived creative agency in robots. We also use the Godspeed scale to explore different attributes of social robots in this setting. We did not obtain significant main effects of the robot movements or the story in the participants' scores. However, we identified significant main effects of the robot movements in features of animacy, likeability, and perceived safety. This initial work encourages further studies experimenting with different robot embodiment and movements to evaluate the perceived creative agency in robots and inform the design of future robots that participate in creative interactions.

Keywords: creative robots, human-robot creative interaction, games, story-telling, StoryCubes, perceived creativity in robots, social robots

## 1 INTRODUCTION

An important dimension in social interaction is how agents perceive the intelligence and creativity of other agents. However, creativity is an under-explored area in the study of Human-Robot Interaction (HRI) (Saunders et al., 2013). Creativity can be defined as the capacity to imagine alternative futures. Despite its relevance in shaping everyday interactions, we have a limited knowledge of how creativity is perceived and attributed by humans to self and others. Furthermore, the display of creativity is a subjective phenomenon that is challenging to study using experimental methods of inquiry (Svanæs, 2013). This study draws from work in the area of self-assessment of creativity to evaluate the possible connections between the perceived creative agency and the animacy and kinaesthetics of social robots.

In artificial agents such as social robots or screen-based avatars, the attribution of creativity has remained largely unaddressed in the field of HRI until recently (Alves-Oliveira et al., 2017, 2020). Whilst Artificial Intelligence has been investigated extensively, Artificial Creativity defined as the creativity attributed to artificial agents remains to be addressed particularly in experimental studies. Some creative behaviors have been simulated using language, pattern recognition, and evolutionary generative systems as shown by (OpenAI, 2021; Gizzi et al., 2019; Pham et al., 2017; Myoo, 2019; DigitalHumans, 2021). However, in artificial agents such as social robots that rely on a physical embodiment to interact with the users and the real-world, artificial creative behavior will need to be communicated to users verbally and kinaesthetically, namely, using movement as part of their communicative means. To our knowledge, the display of creative behavior via physical movement (animacy) in robots is an area requiring further investigation by Human-Robot Interaction researchers.

The display of creative behavior is likely to be of high relevance to determine the ultimate value and usefulness of social robots interacting with humans in everyday settings. The impact of robots' physical presence and their movements has been studied previously across contexts (Vignolo et al., 2017). However, more research is needed to better understand how social robots can effectively use movement in everyday interactions, especially in light of screen-based smart applications and disembodied products that use voice interfaces. What may physical robots offer in terms of functionality and usability that screen and voice agents cannot? And, what are the design affordances enabled by their physicality and movement possibilities that are not available in screen or voice interactions? The work presented in this paper seeks to contribute to the future design of social robots by analysing how humans perceive the robots' movement when these perform a task that requires creative behavior, such as play.

As a novel contribution, we propose a scale to measure the perceived creativity in social robots in the context of games and playful activities such as creative story-telling. This scale is inspired in the work of Karwowski et al. (2018) and Karwowski (2014) and is modified here to capture how participants rate the creative skills of robots. We are particularly focused in the domain of creative collaborative interactions that could be eventually implemented using social robots.

The work with social agents as robots is relevant because excessive screen time is associated with ergonomic, visual and behavioural issues.Furthermore, excessive use of screens for entertainment negatively impacts people at the level of being recently classified as "gaming addiction" (WHO, 2018) and influences people's overall mood, among other effects. At the moment, users intensively use screen-based devices for both work and entertainment services, resulting in extended periods of eye strain and sedentary life-style (Desmurget, 2020; Aboujaoude and Starcevic, 2015; Alter, 2018).

This work is an early exploration of a HRI that does not rely on a screen to interact with users. We aim to study alternatives to screen and audio interactions using physical robots for interactive creative activities relying in more natural interactions with artificial agents. We consider that creativity is a central part of HRI due to its importance in the cognitive and social processes involved in playful interactions. Finally, we expect to contribute in the near future in the design of robots encouraging natural, long-term interactions with cognitive and social gains.

## 2 BACKGROUND

The design of social robots has shown initial evidence of their potential value for usability and functionality to assist users in everyday life. So far, the main applications of notable market success have been to carpet and floor robot cleaners and toys including robotic pets for therapeutic purposes. For the last two decades, researchers and companies have searched for the 'killer app' that builds on the affordances of physical robots to transform the lives of users around the world. In that time, screen-based devices and home assistants that use audio interfaces have made substantial gains in market penetration. Currently, there is a need to understand if and how physical robots will be of value for users in their everyday tasks as suggested by fiction (Miller, 2021).

Arguably the most salient affordance that gives social robots an edge over screen and voice assistants is their physical presence. The importance of physicality and movement in communication is evident from 'body language' to 4E cognition; i.e., the principle that all human cognition is embodied, embedded, enactive, and extended (Lindblom and Alenljung, 2015). In other words, people are not 'brains in jars' but rely heavily on their bodies, the physical world around them, other humans, and their contexts to be able to think, communicate, and be fully human.

There have been studies on issues relevant to the design of robots with movement in mind like the study developed by Hoffman and Ju (2014). In the early stages of research in Human-Robot Interaction researchers as Van Breemen (2004) understood that body gestures are a natural channel to communicate robot's agency and social behaviours. Furthermore, the motion of robot agents (mechanic and organic), as one of the main features differentiating robots from AI or computers has been explored to highlight its relevance from robot aesthetic and functional context by Harris and Sharlin (2011). Similarly, Bainbridge et al. (2011) explored how the physical presence of a robot affects human judgments of the robot as a social partner. Looking for the effects of form and motion in robotic agents Castro-González et al. (2016) studied the attributions of animacy and investigated how the combination of robot bodily appearance and movement can alter attributions of animacy, likability, trustworthiness, and unpleasantness in the users. Apparently, a Baxter robot executing mechanistic movement was perceived as inanimate. However, the same robot performing naturalistic movements was unpleasant. However, all these previous works have not been placed in the context of creative expressiveness of the robot agents.

Our work aims to explore and understand how animacy plays a role in the perception of social robots in the future design of interactive tasks related to creativity. According to the Oxford English Dictionary, *animacy* is "...the quality or condition of being alive or animate" ("Animacy", 2021) whilst *kinaesthesis* refers to "...the effort that accompanies a voluntary motion of the body" ("Kinaesthesis", 2021). In the context of this research, Kinaesthetics and Animacy refers to the study and perception of body motion, and kinetic design refers to the use of movement as a design material (Sosa et al., 2015).

Social and creative games represent interesting settings to study the interaction with social robots as they create a space for playful semi-structured interactions. In many social games, clear rules exist but significant open-endedness is supported to exercise and enjoy the creativity of oneself and others. Creative games with open rules as StoryCubes have not been used often in HRI. However, games are a popular setup in HRI. For instance, (Leite et al., 2009) used chess (to some extent a creative game) as a setup to understand how social presence of robots are perceived. Similarly, interactive storytelling in HRI has been reported as a promising scenario for children's social skills developmentLeite et al. (2015, 2017).

We are interested to examine the physicality and moving affordances of social robots (their animacy qualities), and their potential advantages over traditional board games or interactive games such as mobile apps and voice assistants. It may be possible to combine the best of digital and physical affordances to design social robots that can meaningfully augment social games. To this end, studies are needed to assess the impact of the physical presence and movement of robots in these contexts of use. Story-telling games in particular support creative social interactions. We thus select social games, and particularly story-telling games such as the popular 'Story Cubes' (Ros and Demiris, 2013; Bae et al., 2016; Eladhari et al., 2014; Gordon and Spierling, 2018) as the site of research for this study.

Creativity is an important component of social interaction (Rogers, 1954). Movement has communicative properties that make it an essential part of social interaction between humans (Goldman, 2004) and it is widely regarded as central to embodied experiences including everyday creativity (Svanæs, 2013). Social robots assisting develop creative capacities have been proposed before demonstrating the importance of physical movement in this type of applications (Hoffman and Ju, 2014). Similarly, storytelling is a creative and social activity that has entertainment value but is also used to support learning (Sadik, 2008) and health (Plaisant et al., 2000).

Kinaesthetic creativity has been studied mostly in artistic performance by humans (Tan et al., 2018; Ros and Demiris, 2013). To our knowledge, this work represents an early approach to the study of kinaesthetic creativity in social robots. The research methodology for this study is an experimental design based on previous studies as the ones performed by (Salem et al., 2011; Tung, 2016; Hoffman et al., 2015; Hoffman and Ju, 2014).The effects of robot movement are thus evidenced by their perception of how movement is perceived as a cue indicating creative agency in social robots.

In this study, we chose StoryCubes as an experimental setup. This is a game where players create or re-create stories to share with others, thus requiring to some extent creative skills found universally (Brent, 2014). Our study asks whether social robots will have an edge based on their physicality and their basic animacy features to engage in creative activities such as storytelling games. More specifically, it seeks to assess to what extent may movement make a difference in such settings. If the answer to this question is positive, then further work will be needed to identify and evaluate the kinetic principles for the design of playful and creative social robots more specifically. If the answer is negative, then it is more likely that screens and voice interaction devices will be more adequate and arguably even easier to develop, deploy, maintain, and operate in this type of activities rather than robots that use physical bodies to move and reinforce nonverbal interactions. To this end, our study addresses the research gap in how robot movement is perceived when they perform a creative storytelling activity.

In sum, with studies like this, we aim to contribute to the emergent exploration of Human-Robot Creative Interaction (HRCI). Thus, here we set to identify the hypothesised connection between the perceived creative agency and the animacy of social robots. Our goal is to evaluate the relevance of robot movements to attribute creativity to social robots. At this stage, we aim to provide a benchmark for future experiments using no choreographed robot movements. Similarly, we use storytelling games supported by visual cues to study how movement shapes human's perception of creative agency in robots. We need to highlight as a limitation that most social robots present limitations in terms of dexterity when manipulating objects on a human scale.

## 3 METHOD

### 3.1 Research Goals and Questions

This study aims to assess the connection between perceived creative agency and the animacy of Social Robots. We evaluate the relevance of robot movements to how observers attribute kinaesthetic creativity to social robots. The aim of this experiment is to explore the extent in which robots' movement supports the display of an creative act. We employ a well-known literary tale by the danish author Hans Christian Andersen (Andersen et al., 1995) and a modified version of that story with an alternative ending . The creative act of telling a story associated with the movement could possibly lead to significant different perception by users of the creative agency of robots. This is an exploratory study aiming to respond the following research questions without proposing any hypotheses due to the complex nature of the interaction and the exploratory nature of the study:

1. To what extent do humans perceive creative agency in robots when these tell a story as part of a game (display of a creative act)?
2. To what extent do humans perceive creative agency in robots when these display movements accompanying the delivery of the story?
3. How are robots perceived as creative agents compared with humans?

## 3.2 Materials and Implementation

### 3.2.1 The robot

To evaluate the research questions, we programmed a Robotis Mini Humanoid robot (See Figure 1a) to play our own version of Storycubes in a Wizard of OZ setup. The robot used a female voice in English language (Karen) generated in Mac OS 10.15.7 and played at slow speed (25%) an played using a Bluetooth speaker next to the robot. We chose this robot due its small dimensions for transport and suitability to be used for future experiments using board games. The robot was animated using the Robotis Mini app in iOS 14.4. According to the work of Bernotat et al. (2021) and Kuchenbrandt et al. (2014) robot design can lead to a male or female perception of the social robot. Hence, we decide to use a female voice to neutralise the possible male perception of the robot considering the very sharp and angular design of the Mini Humanoid. The robot movements were presented but not choreographed as they are just a stimulus to indicate robot animacy rather than supporting the delivery of the story. The movements supporting the delivery of the story will be studied in a future experiment.

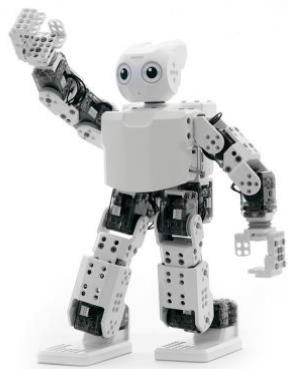
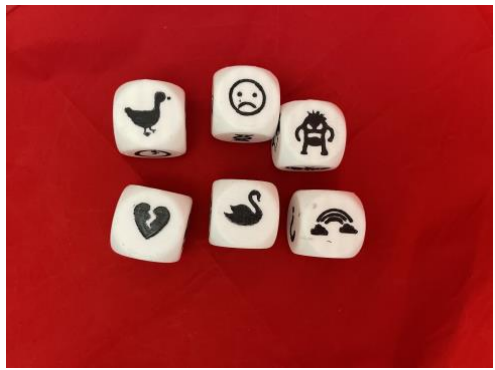

**(1a)** Robotis Mini was the robot used for this experiment due to this small dimensions (Robotis website, 2021).

**(1b)** 3D printed and laser engraved dice designed for this experiment.

**Figure 1.** Robot and StoryCubes used for this experiment.

### 3.2.2 The game

The Storycubes game is a collaborative board games using six or more dice with adaptable rules and an indeterminate number of players. The goal of the game is to create an story with the contribution of all participants. We designed our StoryCubes for this experimentwith a set of six oversized dice that could be visible in the video recordings. After five design iterations using different materials, we used a 30mm, solid-white, PDA, 3D printed dice, laser engraved and hand painted. We modified publicly available icons under the Creative Commons licence for our experiment. Thirty six icons were engraved and six icons were used to tell "The ugly duckling" story for the two different versions used in the experiment.

### 3.2.3 Setup

The standard version of the ugly duckling was split in six short sections to match with the six icons shown in Figure 1b and two additional sections for Introduction and Wrap-up. The creative story was deeply discussed by the authors and validated by four experts in creative writing, English literature, and film scripts in its early and latest versions. The latest version uses dinosaur references which were considered unexpected and perceived as creative without the controversy and the risk of unintentionally offend or hurt feelings of a particular community as the ethics committee suggested (Application HC200985). This last version was validated by one of the specialist in English literature. Both versions, the standard and the creative stories can been read below:

**Original Story:**
*-Ok, Let's start. Once upon a time.* **[Introduction]**
*-There was a mama duck who was very surprised with one of her ducklings.* **[1]**
*-Everyone thought it was very ugly and rejected the poor little duckling.* **[2-3]**
*-She couldn't understand why everyone was so cruel only because she was different.* **[3-4]**
*But a year later, the 'ugly duckling' grew into a beautiful swan!* **[5]**
*-Then she flew away with a flock of swans and every year returned to say hello to her foster mum.* **[6]**
*-She, her mum and siblings celebrate with a happy party around the lake.* **[7]**
*-The moral of the story is that some people take longer to develop and find their true beauty.* **[Wrap-up and Robot Reflection]**

**Creative Story (The bold font indicates the creative twist versus the original story):**
*-Ok, Let's start. Once upon a time.* **[Introduction]**
*-There was a mama duck who was very surprised with one of her ducklings.* **[1]**
*-Everyone thought it was very ugly and rejected the poor little duckling.* **[2-3]**
*-She couldn't understand why everyone was so cruel only because she was different.* **[3-4]**
**-But a year later, the 'ugly duckling' grew into a beautiful flying dinosaur![5.1]**
**-A Pterosaurs called Quetzalcoatlus.[5.2]**
**-She discovered that she was different, not ugly.[5.3]**
**-Then she flew away with a flock of happy Pterosaurs and every year returned to say hello to her foster mum. [5.4]**
*-Then she flew away with a flock of swans and every year returned to say hello to her foster mum.* **[6]**
*-She, her mum and siblings celebrate with a happy party around the lake.* **[7]**
*-The moral of the story is that some people take longer to develop and find their true beauty.* **[Wrap-up and Robot Reflection]**

The robot, the dice, and the speaker (not visible) were allocated in a photo tent in order to isolate them from the external stimuli, avoid distractions for the users and make possible the replication of the experiment. A high contrast, red carpet was used to highlight the dice and the robot. After several tests, we decided to use the video which was recorded in from the top-left corner of the photo tent using an iPhone 12 mini with an 0.8x digital zoom. This setup allows to the viewers to have a full view of the robot and dice. We consider, that using this point of view allows to the user to understand the creation of the story by the robot. See Figure 2. The length of the videos was shorter when the standard story was displayed due to the three extra sentences used in the creative story. These three extra sentences in the creative story aim to reinforce the novel character of the story and assure that the participant noticed the plot-twist. These setup was implemented by the suggestions of the experts in writing and literature.

Four videos were displayed to the participants showing two different robots performing a creative task. The creative task consists of storytelling performed by the robot. The method to tell the story is supported by visual cues in the form of icons on dice and the robot manipulating them. Once the bowl covering the dice was removed, the robot started one of the proposed stories with the movements depending the conditions. The dice and videos of this experiment can be requested contacting the corresponding author. The videos are unlisted in YouTube but can be reviewed using the next links: **Video MS condition, Video MC condition, Video SC condition, Video SS condition.**

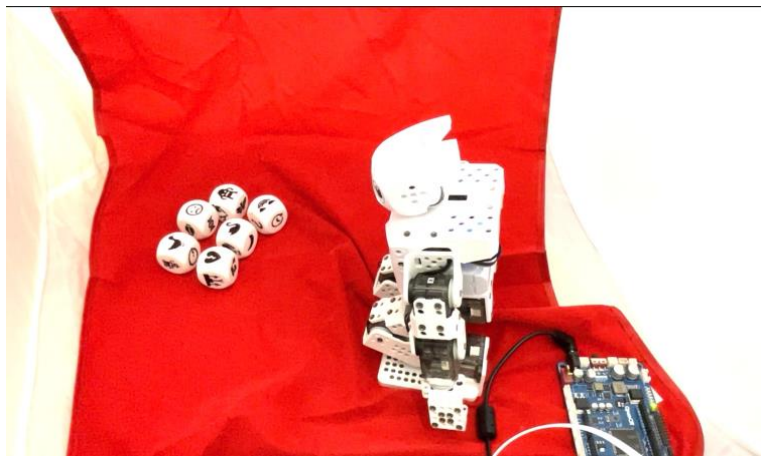

**Figure 2.** Snapshot of the video showed to the participants. The setup used for the four different conditions of the experiment is the same (cubes, robot, and microcontroller in the shot)

In sum, we programmed a humanoid robot to tell the original version of the "The Ugly Duckling" and a modified version of the same story (Creative story) for the experimental condition using cubes with visual imagery such as those used in the "StoryCubes" creative game. In the control condition, the robot remains static while telling the story, while in the experimental condition the robot performs movements to accompany the telling the story kinaesthetically. These interactions were video recorded and participants are requested to fill a survey evaluating the creative agency of the robot.

## 4 EXPERIMENTAL DESIGN

We designed a 2x2 between-subject online study. The factors are the story and the robot movements. The robot can tell the standard story or the creative story (dinosaur plot-twist) and the robot can display movements or not (still) during the story-telling. Hence, we test four conditions display by the robot: still-robot and standard story (SS), still-robot and creative-story (SC), Moving-robot and standard story (MS), and moving-robot and creative story (MC). As a between-subject study, the participants of this online study were expose to one of the conditions mentioned above. I.e, a participant in the condition MC would see a robot gesticulating telling the ugly duckling story with the plot-twist of the dinosaur. See table 1.

| | Experimental Design | |
|---|---|---|
| | Standard story (Control condition) | Creative story (Experimental condition) |
| Still robot | Still-standard condition **(SS)** | Still-creative condition **(SC)** |
| Moving robot | Moving-standard control **(MS)** | Moving-creative condition **(MC)** |

**Table 1.** The four experimental conditions. Each condition shows the factorial combination displaying a particular performance of the robot.

### 4.1 Setup and measurements

This study was approved by the Human Ethics committee of the University of New South Wales, application HC200985 reviewed by the HREAP Executive. Similarly, this study was funded using the Scientia Fellowship (PS46183) development package provided by UNSW. We implemented the survey on Qualtrics licensed for UNSW. A survey flow was created to assign the participants to the four conditions randomly, see Figure 3. The survey was distributed using *Prolific.co*. The survey was designed as follows: Firstly, the participant information and then the consent form. Once the participant agreed to participate in the studio, he/she was directed to the next page and demographic information and confirmation of the Prolific ID was collected. The identity of the participants is anonymous and demographic information was collected such as age, gender, occupation, location, and level of education.

Once the demographic information was collected, the Short Scale of Creative Self (Karwowski et al., 2018) questionnaire was applied. The questionnaire is conformed by eleven questions measuring Creative Self Efficacy (CSE), and Creative Personal Identity (CPI). The questions are as follow:

1. *I think I am a creative person (CPI).*
2. *My creativity is important to who I am (CPI).*
3. *I know I can efficiently solve even complicated problems (CSE) .*
4. *I trust my creative abilities (CSE).*
5. *Compared to my friends, I am distinguished by my imagination and ingenuity (CSE).*
6. *Many times I have proven that I can cope with difficult situations (CSE).*
7. *Being a creative person is important to me (CPI).*
8. *I am sure I can deal with problems requiring creative thinking (CSE)*
9. *I am good at proposing original solutions to problems (CSE)*
10. *Creativity is an important part of myself (CPI).*
11. *Ingenuity is a characteristic which is important to me (CPI).*

Questions one, two, seven, ten, and eleven gauge CPI and questions three, four, five, six, eight, and nine are assigned to the CSE. Following the questions, the participants watched one of the four videos. We confirmed that the video was watched via a check button and requesting a brief description of the video by the participant. Next, the two first authors proposed a modified version of the SSCS to be applied to robots playing creative task. In this case, the task is framed in the playing of our version of StoryCubes. As the original SSCS scale, our Likert scale goes from one to five. The anchors are definitely not (1) and definitely

yes(5). This is a novel contribution proposed by this study that we expect to further develop shortly. The questions proposed are listed below.

1. *I think that the way the robot played StoryCubes shows it is a creative robot.*
2. *The creativity of the robot playing StoryCubes is important to how it behaves.*
3. *The robot efficiently plays StoryCubes.*
4. *I trust the robot's creative abilities to play StoryCubes.*
5. *Compared to other players of StoryCubes, the robot is distinguished by its imagination and ingenuity.*
6. *I think that the robot can consistently create good stories when playing StoryCubes.*
7. *Being creative is important for a player of StoryCubes.*
8. *The robot can deal with problems requiring creative thinking.*
9. *The robot is good at proposing original stories in StoryCubes.*
10. *Creativity is an important part of how the robot plays StoryCubes.*
11. *Ingenuity is a characteristic which is important to how the robot plays StoryCubes.*
12. *I think that the robot can consistently create good stories when playing StoryCubes.* Additional question

Additionally we added a question a twelfth question. The aim of this question is to summarise the general impression of the participant in just one score. The Godspeed questionnaire was applied after the SSCS and general impressions over the study were requested to the participant. The survey concludes confirming the end and providing a code that will allow the compensation for the participant by Prolific. The content of the survey is available by request to the correspondence author.

### 4.2 Participants

A total of 297 participants were recruited using the platform *Prolific*. All participants were over 18 years old and there were no restrictions on gender, formal education, income, or other demographics. Due to technical issues, just 242 participants were exposed to one of the four conditions and compensated by completing the questionnaire. We used the data of 239 participants as the SSCS score of three of them was not recorded. Participants were paid the equivalent of 1.27 British pounds (or 2.2 Australian dollars) for their participation. The first half of the study was run in February 2021, and the second part one week later. 145 participants were male, 89 female, 3 non-binary and 2 did not specify. The average age was 26.9 years old (SD= 8.58). Participants came from a range of locations; 38.5% from North America (Canada, USA, and Mexico), 34.7% from Europe, 13% from United Kingdom, 7.5 % from South America, 4.2% from Oceania, and 2.1% from Africa.The education levels are distributed as follows: 42.7% University degree, 38.5% High School, 9.6% Masters, 1.7% PhD, 5.4% Vocational education, 1.7% Primary education, and 0.4% other. In terms of occupation, 42.3% were students, 38.9% employed (6.3% IT and software, 3.3% artist, 2.5% freelance, 1.3% researcher, and 25.5% other), 12.1% were unemployed, 2.5% homemakers, 0.4% retired, and 1.3% did not specify.

The average time to fill the survey was 9 minutes 20 seconds. We suggested to the participants to use a device with a large screen to have a better user experience and show the video and survey consistently. 72.8% used Windows 10, 8.4% used Macintosh, and 18.8% used other platform (Android 10 5.4%, iPhone 4.2%, Windows 6.1 3.8%, Windows 6.3 1.7%, Android 11 1.3%, Android 9 1.3%, Linux x86 0.8%, Ubuntu 0.4%).

We aimed to allocate at least 60 participants per condition. We did not fully-record data of three subjects lost in the SSCS score. Participants were randomly allocated as follows: (SS)=62 participants (minus two missed SSCS scores), SC=61 participants , MS=61 participants (minus one missed SSCS score) and MC=58 participants. The average Human SSCS score is 3.89 (SD=0.66), no significant differences were found among the different experimental conditions (ANOVA). Participant's results are stored in a standard online spreadsheet and the statistical analysis was made using IBM SPSS.

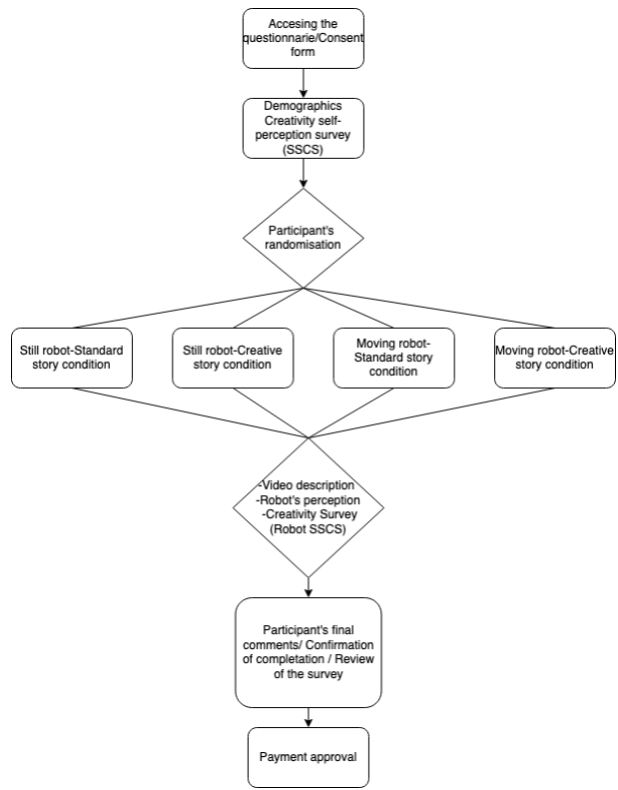

**Figure 3.** Experimental procedure. We applied the questionnaire using Qualtrics. We distributed it using Prolific.co.

## 5 RESULTS

In order to address the exploratory questions of this study, we performed multiple 2x2 factorial analyses of variance (ANOVA), the factors were the story (standard vs. creative) and movements (still, vs moving). We defined seven dependent variables: the SSCS, the question twelve and the five Godspeed items (anthropomorphism, animacy, likeability, perceived intelligence, and perceived safety).In addition, we performed a Pearson's correlation present among the humans and robots CPI, CSE, and SSC scores to check the internal consistency and possible human-robot significant correlations among the scores.

### 5.1 Short Scale of Creative Self (SSCS) score applied to the robots

This variable is our main indicator to assess how participants perceive robots as creative agents. . The average SSCS score of the robots in the four different conditions are as follows: SS= 3.48 (SD=0.74), SC=3.66 (SD=0.61), MS=3.58 (SD=0.68), and MC=3.53 (SD=0.80). As the original SSCS our Likert scale goes from one to five. The anchors are definitely not (1) to definitely yes(5).

We ran a two-way ANOVA with the SSCS score as a dependent variable and the story and movements as factors. Residual analysis was performed to test for the assumptions of the two-way ANOVA. The assumption of homogeneity of variances was not violated, as assessed by Levene's test for equality of variances, p=0.122. Data was normally distributed as assessed by Kolmogorov-Smirnov test (p=0.200).

There were six outliers as assessed as being located less than 3 box-lengths from the edge of the box in a boxplot. They were not removed as a following ANOVA without them showed that they did not affect the results. No significant main effects were found, neither interaction effects. See Figure 4. Similarly pairwise comparisons were run aiming to find simple main effects. However, non significant effects were found again.

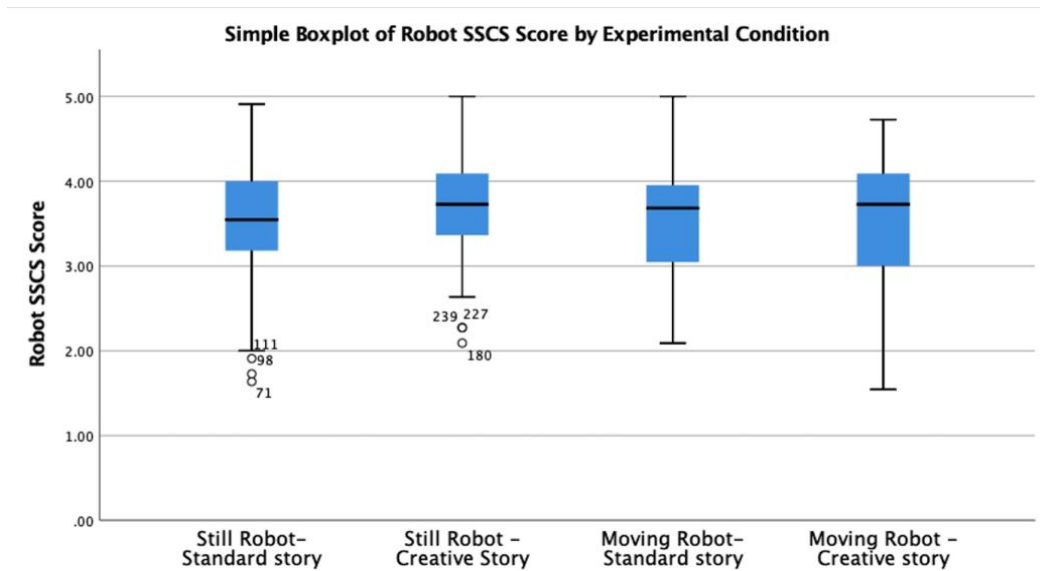

**Figure 4.** Boxplot per group of the SSCS applied to the robot with outliers. No significant differences were found among the means of the groups.

Hence, as suggested in (Laerd, 2021) a further robust ANOVA was run. We used Sigmaplot to run a Kruskal-Wallis non- parametric ANOVA. Once again, no main effect neither interaction effect were found. KW=2.11, df=3, p= .550.

### 5.2  Question twelve and Godspeed items

A similar procedure was followed for the analysis of the rest of the dependent variables. Effects were found only for the items of Animacy, Likeability, and Perceived Safety. Data was normally distributed, as assessed by Kolmogorov-Smirnov test for Animacy (p=0.200), Likeability (p=0.200) but not for Perceived Safety (p <.001) and Q12 (p <.001). The assumption of homogeneity of variances was not violated for Animacy (0.989), Perceived Safety (p=0.173) and Q12 (p=0.909) as assessed by Levene's test for equality of variances. However, it failed for likeability (p=0.006).

We decided to keep the outliers assessed as those being located less than 3 box-lengths from the edge of the box in a boxplot. The results indicate that movement has a main effect in how participants perceive the robots in the different conditions. Animacy $F(3,235)=39.777$, $p <.001$, Likeability $F(3,235)=12.824$, $p <.001$,, and Perceived safety of the robot $F(3,235)=19.127$, $p <.001$. Due to the violation of the assumption of homogeneity for Likeability, a robust non-parametric Kruskal-Wallis ANOVA was run for this variable. Movement has a significant main effect KW = 10.205, df=3, (p=0.017). See Table 2.

### 5.3  Person's correlations among CPI, CSE, SSCS in humans and robots.

A Pearson's 1-tailed correlation was carried out to assess the relationship between the human CPI, CSE, and SSCS scores and similar scores granted to the robot. We aimed to find if correlations between people perceiving themselves as highly creative project this on the robot creative act. The scores of the 239 participants were analysed. There were statistically significant, moderate and strong positive correlations between CPI-CSE, SSCS-CSE, SSCS-CPI in humans, and similar pattern of correlations among the robot scores suggesting consistency in our proposed scale and its internal scores. However, no significant correlation appeared between human scores and robot scores. We should highlight that overall, the means of the human scores were higher than the robot scores in all the experimental conditions and all the scores. See Table 3.

| Means and SD for Question 12 and Goodspeed items for the movement main effects condition. | | | | |
| --- | --- | --- | --- | --- |
| | Still robot-standard story | Still robot-creative story | Moving robot-standard story | Moving robot-creative story |
| Q12 | 3.27(1.06) | 3.25(1.06) | 3.57(0.98) | 3.45(1.08) |
| Anthropomorphism | 2.16(0.73) | 1.99(0.77) | 2.27(0.76) | 2.15(0.76) |
| **Animacy** | **2.36 (0.76)** | **2.33(0.74)** | **3(0.68)** | **2.84(0.66)** |
| **Likeability** | **3.49(0.96)** | **3.61(0.96)** | **3.96(0.62)** | **3.89(0.73)** |
| Perceived Intelligence | 3.48(0.87) | 3.47(0.75) | 3.67(0.70) | 3.45(0.72) |
| **Perceived Safety** | **2.88(0.61)** | **2.74(0.37)** | **3.12(0.54)** | **3.10(0.56)** |

**Table 2.** Bold font means this item had a significant main effect on the movement condition.

| Pearson's Correlations among the CPI, CSE, and SSCS scores for humans and robots ($p < .001$) | | | | | | |
| --- | --- | --- | --- | --- | --- | --- |
| | Human CSE | Human CPI | Human SSCS | Robot CSE | Robot CPI | Robot SSCS |
| Human CSE | 1 | | | | | |
| Human CPI | .573 | 1 | | | | |
| Human SSCS | .883 | .891 | 1 | | | |
| Robot CSE | | | | 1 | | |
| Robot CPI | | | | .673 | 1 | |
| Robot SSCS | | | | .931 | .896 | 1 |

**Table 3.** Pearson's correlations among the CPI, CSE and SSCE scores. The scores are internally consistent.

## 6 DISCUSSION

This study aims to contribute with a benchmark usable for the design of future social robots. With this study, we want to encourage a discussion in the domain of robot's perceived creativity and explore robot movement supporting the delivery of a creative act. However, studying creativity using quantitative approaches presents a number of challenges. We frame our findings in the context of games as a mean to sustain long-term, meaningful human robot creative interactions and several considerations should be taken.

To answer the first research question: *To what extent do humans perceive creative agency in robots when they tell a story as part of a game (display of a creative act)?* We assured that participants watched the video requesting a checkbox validation and a brief description of what they saw. Participants frequently used the word "story" or even specifically "ugly duckling story" in these descriptions. 92.5% of the participants mentioned the word "story" or similar (tale, history, story line, fable) when describing the video. In few cases typos were present and derivations such as history, story or study were used but the intention is taken as the same. Furthermore, 100% of the participants referred directly or indirectly to the act of telling a story even when they did not use the word in their descriptions. For instance; *"It's about a robot that tells ugly duck that turns into swan."*, *"the process to find the real beauty"* or *"It's about a robot that tells ugly duck that turns into swan."*.

We highlight the fact that the SSCS scores of the robots, even when they are not significantly different, were above the mean (2.5) of the 1-5 scale of the SSCS score as indicated in section 5.1. Even though the robot scores are lower that the human scores, the minimal score was above 3.4 for the still robot-standard story condition. The scores per condition were: SS= 3.48 (SD=0.73), SC=3.66 (SD=0.61), MS=3.58 (SD=0.68), and MC=3.53 (SD=0.80). Hence, we can claim that participants were aware that the robot was performing a creative act delivering a story. Future work can involve further statistical analysis comparing with an specific benchmark indicating what is a minimal score indicating that a social agent is considered a creative agent. Similarly, a face-to-face setup could be more appropriate to perform an experiment of this nature since it is likely that the robot embodiment has a significant impact in the participant's perceptions compared to virtual agents.

For the second question: *To what extent do humans perceive creative agency in robots when these display movements accompanying the delivery of the story?* We considered that robot movement would be a variable moderating the effect of the story in how people perceive robots. The marginal means graph can wrongly lead to conclude that movements moderate the SSCS score. However, when inspecting the boxplot, it is clear that means among all the experimental conditions are not significant. See 4. Although we did not notice main or interaction effects in the SSCS score, we did notice significant effects in three of the items of the Godspeed scale. These items are animacy, likeability, and perceived safety.

In the case of animacy, we observed that participants in this study could notice the movements of the robot as they score significantly highly in animacy to the moving robots independently of which story the robot is telling. This shows that participants are aware of the movement and how the movement impacts participant's perceptions in terms of likeability as they rank moving robots MS=3(0.68) and MC=2.84(0.66) significantly higher than still robots SS= 2.36(0.76) and SC= 2.33(0.74). See Table 2.

The robot movements were not mentioned frequently in the description of the video as the story. However, some participants used anthropomorphic terms. I.e. *"A creepy robot gives it's version of the ugly duckling by Hans Christian Andersen. It also does a decent MC hammer impression.""...The movements of the robot were however a bit erratic and didn't match that well with the story it was telling." "In a way I feel like I was arranged to tell that story but I liked the movements and the appearance of the robot"*.

In terms of likeability, as shown in Table 2, participants scored significantly higher to the moving robots in terms of likeability. MS=3.96(0.62) and MC=3.096(0.62) significantly higher than still robots SS= 3.49(0.96) and SC= 3.61(0.96). The significant main effect of the movements in likeability is aligned with previous studies using games (Sandoval et al., 2016b,a, 2020). Apparently, humanoid robots tend to be likeable when they perform unexpected tasks that can be interpreted as social or creative. Future studies could test other robot embodiment perceived as less humanoid. An illustrative comment in how some participants perceive the robot was: *"How this robot looks like and how it feels like compared to a human being. By his movements, he looked really happy, by only talking sometimes we cant understand how smart a robot can be. He was very smart and we can definitely see that his voice wasn't recorded at some point."*

At the beginning of this experiment we considered that the factor of movement would be a moderating variable supporting the delivering of the story by the robot. In other words, robot movement would lead to higher scores for the robots in both kind of stories or at least in one of the stories. However, evidence for this was not registered, a possible reason for that is the type of the robot's movement. For this experiment, we intentionally design a set of robot movements that are were part of the standard programming of the robot but are not synchronised with the delivery of the story. The reason for this is that in future applications using robots for social games, it is unlikely that robot movements will always be customised to their dialogues. Better synchronised and choreographed movements would be an obvious stimulus that could cause significant main effects in the SSCS score and how people perceive robots as creative agents. This type of movements can be tested in future studies. In terms of perceived safety, moving robots were perceived as safer than still robots. This is a result worth considering further, especially taking into account that this study is an online experiment and not a face-to-face setup.

Question twelve in the survey: *"I think that the robot can consistently create good stories when playing StoryCubes"* was added to allow the participants to summarise their impressions from the previous SSCS questions. The movement main effect was not significant ($p = 0.64$), when the participants answered this question. However, this provides an intriguing result to be explored in future studies as the comments of the participants suggest. Certainly, the participants perceived the movement of the robot but not necessarily the novelty of the story. Independently of the story, participants rated slightly higher the moving robots compared with still robots as agents which can create good stories when playing StoryCubes. Even as a marginal result, this finding points to the importance to study a range of robot movement approaches in future work. Further qualitative analysis are required to explore.

For question three in the survey, *How are robots perceived as creative agents compared with humans?* in all the experimental conditions robots were score lower than humans for the SSCS scores. Robot scores can be seen in Section 5.1 and human SSCS scores are as follow SS= 3.88(SD=0.68), SC=3.85 (SD=0.83), MS=3.88 (SD=0.70), and MC=3.95 (SD=0.66). Pearson correlation does not suggest any significant correlation among the human and robot scores. However, there are significant correlations among the sub-scores CPI and SCE in both human and robot scores which could suggest that the internal consistency of the original scale and our adaptation for assessing robot creativity have potential as measurement tools for future studies in Human-Robot Creative Interactions. See Table 3.

Some of the participants' comments suggest that a further exploration of this question could be relevant as they were enthusiastic about the capabilities of the robots comparing with previous human creative experiences. To illustrate, *First of all, I adored the robot. I thought it was cute. Plus I'm impressed with it's abilities, too! I mean I can probably kinda guess how it works, but still - it's just mindblowing! And I love that the message of the story told by the robot is so wholesome! As a musician and somewhat a songwriter, I find it astonishing how it can come up with a good story in such a quick amount of time and keep it up. As for the study itself, I like the way the text light up when the mouse hovers over it, I haven't seen it a lot, if anytime. In addition, i had to check up on two of the English words used in the study, which i highly value as an educational feature. I thoroughly enjoyed the experience.* And *"This was an interesting concept to consider, and I'd honestly like to see more content involving AI and StoryCubes."* and *Im impressed how robot can told us a story based on random images.*

## 7 CONCLUSIONS

Creativity can be considered an aspect of autonomy and agency in social agents that is different from intelligence, logic, and strategy. The current understanding on how creativity is displayed by robots is still limited. This study aimed to inform the design possibilities of Human-Robot Creative Interaction (HRCI) and provides a reference for future studies exploring the main factors involved in the creative interaction between humans and robots. Our findings show that the setup used in this study does not trigger higher scores in the SSCS scale differentiating the robots as creative agents. However, movement does shows main effects in the scores of animacy, likeability, and perceived safety of the Godspeed scale. Further, the scores of the moving robots were above the media in all the cases (although they were lower than the SSCS scores in humans). In terms of how robots are perceived as good story-tellers (question 12 in the robot

SSCS), even when the scores are not significantly different, the results provide an important insight in how to continue the development of future experimental studies. For instance, the need of perform similar study in a face-to-face setup and using other robot embodiments beyond humanoid robots.

The study of creativity in robots shows a research gap when addressed in playful, creative, and collaborative activities such as board games. We chose a playful task (a storytelling game) to empirically evaluate the extent to which a robot's physical embodiment may cause humans to attribute creative agency to a robot. The StoryCubes game offers a means to further assess the display of creativity in robots considering the applicability of this setup as entertainment and for the development of cognitive skills, spacial memory, decision making and collaborative skills (Unbehaun et al., 2019; Wu et al., 2012).

Furthermore, we consider that our approach is useful aiming strategies for long-term interaction and as an alternative to avoid screen addiction and contribute to a better mental health in the digital age (Sandoval, 2019; Aboujaoude and Starcevic, 2015). Even in the StoryCubes mobile app, the user experience is visibly compromised when compared to the physical cubes. Considering this, we highlight the importance of perceived creativity in social robots to further develop the early work in Artificial Creativity. It looks like it is critical to explore advantages and disadvantages among robotic interfaces that display and support creative interactions. To this end, when users are exposed to stimuli related to creative robots it seems critical to set their expectations in this type of studies. One participant said, for instance: *"The study itself was fine. The premise, however, is something that hasn't been fully explained. For example, is this a prototype of a children's toy? Is it a learning device? Is it a diagnostic tool?"*. Finally, one of the comments of the participants is encouraging to continue the development of studies in creative robots playing story-telling games: *"Assuming the robot really came up with the story using its own creativity and it wasn't programmed into it, I am very impressed with the level of depth which the story had. In that sense, the robot could even be more creative that a lot of humans. I also think its use of words and its manner of speech is properly human-like. That is not to say you couldn't feel the robotic nature behind it at all. In my opinion, some work should be done on the robot's movement and how it connects to whatever it is saying in a way that makes more sense. I wish you good luck with the study and with further development of creative robots."*

## 7.1 Limitations and Future Work

The main technical limitations to implement a robot board game have been discussed before by (Sandoval et al., 2021). The game StoryCubes in particular has a significant element of improvisation and randomness, and the translation of the cubes to create a consistent story is a technical challenge for robots that genuinely synthesize stories from these stimuli. Similarly, the vision system required to read the cubes accurately under a range of lighting conditions and angles would require significant technical work. We are currently working on an implementation where participants and robots play StoryCubes in a shared physical setup for a future study. Choreographic movements of the robots will be programmed and displayed on a bottom-up strategy that starts by incorporating movements in an increasing level of sophistication and detail. We consider that this would inform robot designers to incorporate movement that achieves a balance between creating a meaningful Human-Robot Creative Interaction (HRCI) while drawing from a library of gestures, postures, and body movements suitable for fluid communication. In terms of data collection and analysis, we plan to conduct a thematic analysis of participants' comments. Furthermore, future experimental designs could include different robot embodiments (humanoid versus non-humanoid), variants of StoryCube games, different stories (original stories vs well-know stories). Finally, a more exhaustive validation of our version of the SSCS scale (performing factorial and reliability analysis) may be required to assess future interaction face-to-face in a more robust manner.

## 8 CONFLICT OF INTEREST STATEMENT

The authors declare that the research was conducted in the absence of any commercial or financial relationships that could be construed as a potential conflict of interest.

## AUTHOR CONTRIBUTIONS

Dr. Sandoval's contribution: This author share first authorship: Literature review, conceptualisation of the study, experimental design (conceptualisation of metrics and implementation), ethics approval, data collection, data analysis, writing of the first draft, edition of the manuscript, formatting of the draft in latex.

Dr. Sosa's contribution: This author share senior authorship. Literature review, conceptualisation of the study, experimental design (conceptualisation of metrics), ethics approval, edition of manuscript, proof-reading.

Prof. Cappuccio: This author share senior authorship. Feedback, edition of the manuscript. Prof. Bednarz: This author share senior authorship.Feedback, edition of the manuscript.

## FUNDING

This project was funded by the UNSW Scientia Fellowship PS46183.

## ACKNOWLEDGMENTS

Dr Sandoval wants to thank to Prof Sosa for all the intense and highly collaborative discussions that lead to this paper and the rest of the authors for their valuable contributions. Also thanks to Prof Alfredo Jimenez (IPN-CEPROBI) for his advice on selecting and running additional statistical analysis. Thanks to Karam Hussain at the UNSW Makerspace Paddington campus for his support in the development of the StoryCubes used in the experiments. Also thanks to the story reviewers Carolina Posadas, Jennifer Breakelaar, and Miranda Verswijvelen. Dr. Sandoval wants to thank to Sebastian Sandoval-Garcia to inspire him to finish the first draft of this manuscript one day before his birth.

## SUPPLEMENTAL DATA

Link in UNSW Onedrive to access the data files can be found **here**.

## DATA AVAILABILITY STATEMENT

The dataset and analyses of this study can be found in the UNSW Onedrive **here**.